# Explaining the Unexplained: Revealing Hidden Correlations for Better Interpretability

Wen-Dong Jiang [a,1], Chih-Yung Chang [a, 2, *], Show-Jane Yen [b, 3] and Diptendu Sinha Roy [c, 4]

[a] *Department of Computer Science and Information Engineering, Tamkang University, Taoyuan City 333, Taiwan.*

[b] *Department of Computer Science and Information Engineering, Ming Chuan University, New Taipei 25137, Taiwan.*

[c] *The Department of Computer Science and Engineering, National Institute of Technology, Shillong, 793003, India.*

**Abstract:** Deep learning has achieved remarkable success in processing and managing unstructured data. However, its "black box" nature imposes significant limitations, particularly in sensitive application domains. While existing interpretable machine learning methods address some of these issues, they often fail to adequately consider feature correlations and provide insufficient evaluation of model decision paths. To overcome these challenges, this paper introduces Real Explainer (RealExp), an interpretability computation method that decouples the Shapley Value into individual feature importance and feature correlation importance. By incorporating feature similarity computations, RealExp enhances interpretability by precisely quantifying both individual feature contributions and their interactions, leading to more reliable and nuanced explanations. Additionally, this paper proposes a novel interpretability evaluation criterion focused on elucidating the decision paths of deep learning models, going beyond traditional accuracy-based metrics. Experimental validations on two unstructured data tasks—image classification and text sentiment analysis—demonstrate that RealExp significantly outperforms existing methods in interpretability. Case studies further illustrate its practical value: in image classification, RealExp aids in selecting suitable pre-trained models for specific tasks from an interpretability perspective; in text classification, it enables the optimization of models and approximates the performance of a fine-tuned GPT-Ada model using traditional bag-of-words approaches.

Keywords： Deep Learning, Explained Artificial Intelligence (XAI), Machine Learning, Interpretability.

## 1. Introduction

Thanks to the rapid advancement of computer hardware, deep learning has made significant progress in the application of unstructured data, such as images (Cao & Chen, 2025) and text (Li et al., 2024). Specifically, the success of representation learning (Wang & Lian, 2025; Zhang et al., 2025) has gradually replaced the earlier approaches of transforming unstructured data into structured formats. The key to the success of representation learning lies in leveraging a large number of parameters for backpropagation, enabling the model to adapt to data with non-normal distributions. Although models based on backpropagation neural networks (Yang et al., 2019; Banerjee et al., 2023) have achieved significant technical advancements, their application in many sensitive domains, such as medicine (Zhang et al., 2025) and industrial inspection (Rathee et al., 2021), still faces considerable challenges due to the difficulty in understanding the basis of their decision-making.

Explainable Artificial Intelligence (XAI) aims to reveal the inner mechanisms of neural network decisions, thereby making these models more reliable for applications in sensitive domains. In recent years, several studies (Li et al., 2025; Jing et al., 2025; Liu et al., 2024; Guan et al., 2024) have focused on injecting explainability into deep learning models and using various visualization techniques to explain the decisions of these "black box" models. While these models have achieved a certain level of interpretability, two pressing issues remain (Huang & Marques, 2023; Huang & Marques, 2024): first, whether the correlations between different attributes are correctly evaluated, and second, whether the model's decision-making pathway truly aligns with human reasoning, even when the model's understanding appears consistent with user expectations.

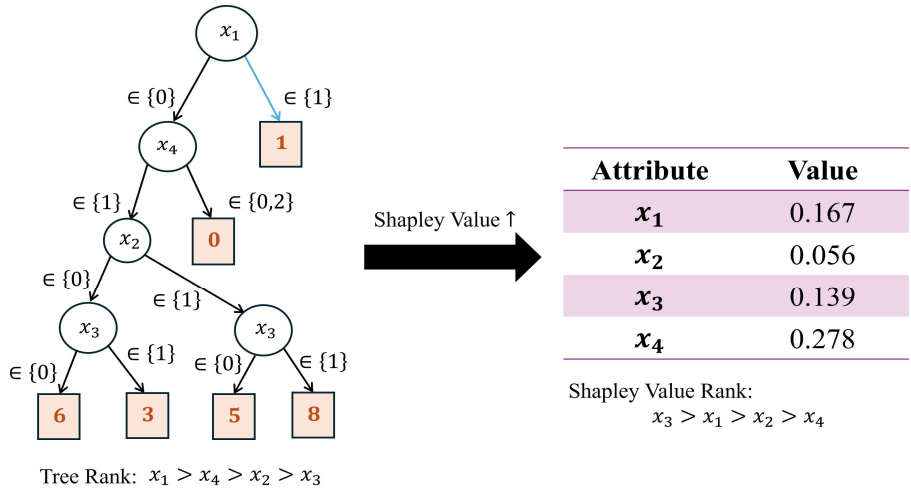

**Fig.1.** Feature Correlation Issues in Explainable Machine Learning

The first issue is that feature correlation was not sufficiently considered. Fig. 1 provides an example: as shown in Fig. 1, the left side depicts a binary tree with four nodes, $\{x_1\}, \{x_2\}, \{x_3\}$ and $\{x_4\}$, each representing a feature. The values on the edges, such as "∈ {0}", "∈ {1}", and "∈ {0,2}", indicate the branching directions determined by the condition of each feature. The values at the leaf nodes (e.g. "6"、 "3"、 "5"、 "8") are the predicted outcomes of the model along that path.

By observing the decision tree, the importance of each feature can be inferred: features that influence more splits and are closer to the root node have a greater impact on the overall decision. In this tree, feature $x_1$ is at the root node, directly deciding which child node to proceed to, thus indicating that $x_1$ is the most important feature in this tree. Overall, the feature importance ranking can be represented as: $x_1 > x_4 > x_2 > x_3$.

However, when using explainable machine learning methods such as Shapley Value (Lundberg & Lee, 2017) (higher Shapley Values indicate higher importance) to compute feature importance, an incorrect result was obtained, with the rank being $x_3 > x_1 > x_2 > x_4$. There are two main reasons for this discrepancy:

First, Shapley Value does not sufficiently account for feature correlation during calculation. It primarily considers the interaction between features without considering the traditional individual importance of each feature. Secondly, when there is redundancy or strong collinearity between features, SHAP distributes the contribution among them, leading to lower Shapley Values for some features, even if they occupy important positions in the decision tree. The specific proof of this issue will be described in detail in **Section 3.1**.

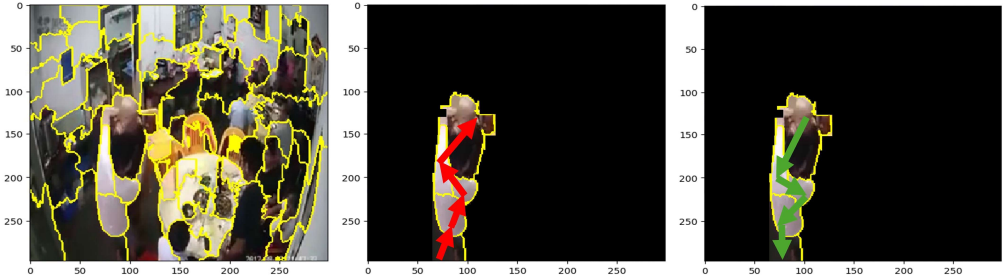

**Fig.2.** Comparison of Model Decision Path and Expert Decision Path for Violent Scene Identification

The second issue is that existing explainable methods often only consider whether the model's decision outcome is consistent with that of humans, without considering whether the decision-making path of the model matches that of human reasoning. Fig. 2 shows a simple example, where from left to right are the original image, the decision path as perceived by the model (Jiang et al., 2024), and the decision path as perceived by the expert. As shown in Fig.2, the areas identified as violent by the model are consistent with the expert's assessment, but the reason behind the model's decision involves a sequence from black pants to the white shirt, and then to the fighting action. This reasoning is precisely opposite to the expert's view of the violent region.

The goal of explainable methods is to provide domain experts with a reasonable basis for deploying models in high-risk scenarios. Therefore, even if the areas identified by the model match those identified by the expert, or even match perfectly at the pixel level, it may still be insufficient to establish trust for high-risk tasks. Clearly, such an explanation is inadequate.

To address the aforementioned issues, this paper proposes a method for interpretability computation called Real Explainer (RealExp). Specifically, this study revisits the calculation of Shapley Values, decoupling the original Shapley Value's average contribution into independent contributions and interaction contributions. Additionally, an interpretability evaluation approach combining expert annotations and the tau coefficient is proposed to further assess deep learning models.

In practice, the proposed RealExp is designed as an interpretable approach for surrogate models, making it capable of explaining any deep learning model. For example, in Image classification task, given an image that requires interpretation and a model, the RealExp masks certain regions of the image at a fixed ratio to generate a series of perturbed images. These perturbed images are then fed into the model to calculate changes in the model's predictions. Subsequently, RealExp fits the output of these new samples using an ensemble tree model, thereby efficiently approximating the marginal contribution of each feature. To ensure the accuracy of the fitted model results, RealExp utilizes an exponential kernel function to fine-tune the weight given to different feature combinations, where the weight is determined by the similarity between the perturbed samples and the original sample. The higher the similarity, the greater the weight assigned. The feature importance scores obtained from the fitted tree model represent the final feature attributions, which reflect each feature's contribution to the model's prediction.

This study extends previous research (Lee et al., 2024) by "Not Just Explain, But Explain Well: Interpretable Machine Learning Based on Ensemble Trees." The primary contributions of this paper can be summarized in three key aspects:

**1. Proposing a New Explainability Computation Method:** This paper introduces the RealExp method to address the lack of trust in existing explainability methods in high-risk scenarios. RealExp revisits the computation of Shapley Values by decoupling the average contribution into independent contributions and interaction contributions, thereby refining the understanding of feature contributions.

**2. Designing an Explainability Method for Surrogate Models:** RealExp serves as an interpretability tool for surrogate models and can explain any deep learning model. Specifically, RealExp generates perturbed images and uses an ensemble tree model to fit the output, calculating the marginal contribution of each feature to more efficiently and accurately assess the model's dependence on different features.

**3. Evaluation Method Combining Expert Annotations and Tau Coefficient:** To further evaluate the interpretability of deep learning models, this paper proposes an evaluation method that combines expert annotations with the tau coefficient. This evaluation method aims to enhance the reliability of explanations, ensuring that the model interpretations align better with domain experts' needs, especially to increase expert trust in the model in high-risk scenarios.

The remainder of the paper is organized as follows. Section 2 discusses and compares previous relevant studies. Section 3 details the proposed RealExp. Section 4 describes the assumptions and problem formulation in detail. Section 5 provides experiments and performance evaluation. The conclusions are discussed in Section 6.

## 2. Related Work

This section will review research works related to Interpretable machine learning, divided into three parts: the first part focuses on perturbation-based explainable methods; the second part covers gradient-based visualization methods for interpretability; and the third part discusses embedding-based interpretability methods.

### 2.1. Perturbation-Based Explainable Methods

These methods explain models by randomly modifying parts of a single test data entry and re-inputting the modified data into the model to obtain new predictions. By comparing the changes in prediction values, the importance of features in the test data could be estimated. The core idea of these methods was to generate perturbed data and observe the sensitivity of the model's output to input changes, thereby inferring the importance of input features.

For example, (Zeiler & Fergus, 2014) slid a gray box window across images and observed changes in the model's predictions to identify the regions the model considered most important. This method intuitively revealed the areas of input data that the model focused on by occluding different regions. (Ribeiro et al., 2016) randomly masked specific regions of images or text to generate new predictions and used the slopes of linear regression models constructed from these predictions to explain the decision-making process of the model. This method was applicable not only to visual data but also to text and other data types. (Petsiuk et al., 2018) further extended this approach by generating inputs with random masks to estimate feature importance, emphasizing the broad applicability of perturbation-based methods in feature evaluation.

Additionally, (Ancona et al., 2019) proposed an approximate explanation method based on Shapley Values to assess feature importance in deep learning models. This method adopted the concept of cooperative game theory to provide a mathematically rigorous explanation framework. (Lundberg & Lee, 2017) suggested quantifying differences between random samples and original samples by calculating the probability derivatives of Shapley Values to explore the model's decision logic. (Zafer & Khan, 2021) used clustering methods to generate perturbed samples, enhancing the interpretability of models. (Liu et al., 2024) combined interpretable tree structures with SHAP value analysis to provide more structured explanations from a tree-based model perspective. (David et al., 2024) developed the AcME method, which optimized the computational efficiency of SHAP values, significantly improving the speed and practicality of the explanation process. (Li et al.,2023) proposed $\mathcal{G}$LIME, which combined an improved Elastic-Net estimator to refine local explanation results by balancing global explanation distances and sparsity/feature selection in the explanations. (Tan et al.,2024) introduced GLIME with a local and unbiased sampling distribution to generate explanations with higher local fidelity.

Although these mask perturbation-based methods are applicable to any black-box model and demonstrate good flexibility in feature importance analysis, two main issues remain: 1. Limitations of the Proxy Model: When these methods rely on proxy models, such as linear regression, for explanations, the inherent flaws of the proxy model can affect the accuracy and applicability of the explanations. For example: Limitations of the Linear Assumption: Linear models (e.g., linear regression) assume a linear relationship between input features and output, while real-world data and models often exhibit complex nonlinear characteristics. Such linear assumptions may fail to capture the complexity of the model's actual decision-making process, thereby reducing the reliability of the explanation. Feature Collinearity Issues: When there is collinearity among input features, linear models struggle to distinguish the independent contributions of each feature, which can lead to bias in feature importance assessments, making the explanations unable to accurately reflect the model's actual behavior.

Additionally, the applicability of these methods is constrained by certain additional factors. For instance, in the case of high-dimensional data, the masking or occlusion process may introduce extra noise, making the results more difficult to interpret.

## 2.2. Gradient-Based Visualization Methods

The core purpose of these methods is to determine which input features have a significant impact on the model's output through one or more backpropagation processes. By analyzing the gradient information of the model, these methods attempt to reveal which parts of the input data are most important to the model's prediction.

For example, (Zhang et al., 2018) used the backpropagation process to trace the activation input pixels of specific nodes or classes within the network, in order to explain the model's decision-making process. This method intuitively demonstrates the correlation between input features and the network structure; (Selvaraju et al., 2017) introduced the Grad-CAM method, which generates visual heatmaps by leveraging gradient information from convolutional layers, highlighting regions crucial to the model's prediction, thus allowing users to intuitively understand the key parts the model focuses on; (Binder et al., 2016) proposed a technique that proportionally backpropagates output errors to each input feature, quantifying the impact of each feature on the prediction and representing it as feature importance; (Shrikumar et al., 2017) assigned "importance scores" to each feature by comparing the contribution of each input feature with that of a reference input, thereby further improving the accuracy of interpretations; (Smilkov et al., 2017) smoothed the gradient maps by adding random noise to the input image and calculating the average gradient, making the generated visualizations more stable and clearer; (Zahavy et al., 2016) highlighted important pixels in the input data by generating Jacobian Saliency Maps, revealing which regions have the greatest influence on the model's value or behavior predictions; (Bakchy et al., 2024) further developed gradient-based interpretability methods by proposing a lightweight CNN-based model, integrating the use of Grad-CAM to enhance usability through fast computation and efficient interpretation; (Niaz et al., 2024) proposed the Increment-CAM method, which combines Grad-CAM computation, Score-CAM computation with a modified approach, heatmap merging, and regularization to achieve model interpretability; (Li et al., 2024) proposed the Contrast-Ranking Class Activation Mapping (CR-CAM), for use with CNNs and GCNs to generate class activation maps. To address the similarity between different categories, the ranking block adopts a comparative approach to measure the distances between feature mappings in manifold space, thereby reducing the weights of the surrounding regions.

Although these methods provide intuitive and effective visual tools to explain deep learning models and reveal the internal mechanisms of model decisions, they still face three main challenges: The first is Difficulty in Providing Clear Explanations for Models with a Large Number of Parameters: As the scale of deep learning models continues to expand, the number of parameters grows exponentially. This complexity can cause gradient information to become dispersed or blurred during propagation, making it difficult to clarify the contribution of a single input feature. For example, in deep convolutional neural networks, as the number of layers increases, issues like vanishing gradients or exploding gradients may occur, thereby reducing the accuracy of backpropagation-based explanations. Additionally, parameter redundancy and model overfitting may interfere with the interpretative process, making it challenging to generate feature importance maps that have practical significance. Secondly, Mainly Applicable to Convolutional Neural Networks or Graph Neural Networks, Limited to Image Data: These methods are mostly designed for CNN and their variants, which perform well in interpreting image data. However, for other types of data, such as time-series data, text data, or tabular data, gradient-based methods are often difficult to apply directly. For example, in natural language processing tasks, the inputs are typically discrete words or characters, which makes it challenging to use gradient visualizations to clearly identify the model's focus on a particular input feature. Furthermore, for tabular or structured data, gradient methods lack adaptability and flexibility, making it difficult to provide effective interpretations. Lastly, Inability to Provide Global Explanations for the Dataset: Gradient-based methods typically focus on providing local explanations for specific data instances, helping to understand the model's decision-making process for a particular input. However, these methods struggle to provide a global explanation for the entire dataset. For example, they cannot reveal differences in model performance across different data subsets, nor can they summarize the global patterns captured by the model. This limitation is significant for applications that require a comprehensive understanding of model behavior, such as medical diagnosis and financial decision-making.

## 2.3. Embedding-Based Visualization Methods

These methods enhance data interpretability through nonlinear dimensionality reduction techniques. The basic idea is to project the complex structure of high-dimensional data into a lower-dimensional space, making the internal relationships between models or data easier for humans to understand. These techniques are typically based on similarity in high-dimensional space, mapping data points to nearby locations in the lower-dimensional space to reveal inherent patterns or distribution characteristics in the data.

For example, (Mnih et al. ,2015) demonstrated how to use t-SNE, a dimensionality reduction tool, to visualize complex high-dimensional data as points on a two-dimensional plane. This method minimizes the divergence between the probability distribution of high-dimensional data points and the probability distribution of the low-dimensional mapped points, allowing the distribution of data in two dimensions to reflect its original structure. This intuitive visualization makes it possible to analyze the internal relationships of high-dimensional data, such as the distribution in clustering or classification. (Engel & Mannor ,2001) proposed an embedding-based mapping method that measures the similarity between data points using transition probabilities. The embedding map generated by this method can effectively show the relationships among data points in a specific feature space, providing a new perspective for interpreting model behavior. (Bibal et al.,2023) proposed DT-SNE, an improved version of t-SNE based on a decision tree model. DT-SNE integrates structural information from decision trees, resulting in dimensionality reduction that better conforms to the original data distribution and characteristics. Furthermore, this approach adds constraints to interpretative tasks, providing more interpretable results compared to traditional t-SNE.

These methods effectively utilize statistical means for interpretation but face challenges when dealing with large-scale datasets, primarily involving the difficulty of designing and selecting appropriate features. Additionally, embedding methods generally preserve only a portion of the properties of the original n-dimensional data, making the interpretation process challenging. Moreover, t-SNE is inherently an unsupervised method, lacking the ability to ensure that instances of the same class remain similar in the visualized result.

**Table 1**

The comparison between Related Works.

| Related Works | Methods | Generality | Attribute Correlation | Paths |
|---|---|---|---|---|
| Zeiler & Fergus, 2014 | Perturbation | ✓ | ✗ | ✗ |
| Ribeiro et al., 2016 | Perturbation | ✓ | ✗ | ✗ |
| Petsiuk et al., 2018 | Perturbation | ✓ | ⚹ | ✗ |
| Ancona et al., 2019 | Perturbation | ✓ | ⚹ | ✗ |
| Lundberg & Lee, 2017 | Perturbation | ✓ | ⚹ | ✗ |
| Zafer & Khan, 2021 | Perturbation | ✗ | ✗ | ✗ |
| Liu et al., 2024 | Perturbation | ✓ | ✗ | ✗ |
| David et al., 2024 | Perturbation | ✓ | ✗ | ✗ |
| Li et al.,2023 | Perturbation | ✓ | ✗ | ✗ |
| Tan et al.,2024 | Perturbation | ✓ | ✗ | ✗ |
| Zhang et al., 2018 | Gradient | ✗ | ✗ | ⚹ |
| Selvaraju et al., 2017 | Gradient | ✗ | ✗ | ⚹ |
| Binder et al., 2016 | Gradient | ✗ | ✗ | ⚹ |
| Bakchy et al.,2024 | Gradient | ✗ | ✗ | ⚹ |
| Niaz et al., 2024 | Gradient | ✗ | ✗ | ⚹ |
| Li et al., 2024 | Gradient | ✗ | ⚹ | ⚹ |
| Engel & Mannor ,2001 | Embedding | ✗ | ⚹ | ⚹ |
| Mnih et al. ,2015 | Embedding | ✗ | ⚹ | ⚹ |
| Bibal et al.,2023 | Embedding | ✗ | ⚹ | ⚹ |
| RealEXP (Ours) | Perturbation | ✓ | ✓ | ✓ |

Table 1 summarizes the comparison between previous research findings and the proposed RealExp, analyzed from four perspectives: Generality, Attribute Correlation, and Evaluation of Explanation Paths. In the table, "✓" indicates that the feature is present, "✗" indicates that it is not, and "⚹" indicates that the feature is present but has certain limitations.

## 3. The Proposed RealExp

This section will introduce the specific details of the proposed RealExp, which are divided into three parts: 1. The background of the Shapley Value and a description of the existing problems. 2. The proposed RealExp and the corresponding proofs. 3. The steps for implementing RealExp in practice. The specific details are as follows:

## 3.1 Shapley Value

The Shapley Value is a fair method used in game theory to distribute cooperative gains. It determines each participant's contribution to the overall profit by calculating their marginal contributions in different combinations. The Shapley Value has desirable properties such as symmetry, efficiency, and gain

independence, which ensure fairness in the allocation process. The Shapley Value is given by Exp. (1):

$$\Delta_v(i, S) = v(S \cup \{i\}) - v(S)$$

$$\varsigma(|S|) = \frac{|S|!(|\mathcal{F}|-|S|-1)!}{|\mathcal{F}|!} \tag{1}$$

$$Sv(i) = \sum_{S \subseteq (\mathcal{F}\setminus\{i\})} \varsigma(|S|) \times \Delta_v(i, S),$$

The core idea of the equation is to measure the marginal contribution of feature $i$ in different feature subsets $S$ for the model output $v$. Herein, $\Delta_v(i, S)$ represents the marginal contribution of feature $i$ to set $S$. $v(S)$ denotes the profit generated by set $S$ in model $v$, while $v(S \cup \{i\})$ represents the profit generated after adding feature $i$ to set $S$. $\varsigma(|S|)$ is a factor representing the weights of different subset combinations in the calculation. Essentially, the Shapley Value is the average of marginal contributions, which can also be represented by Exp. (2):

$$Sv(i) = \frac{1}{|F|!} \sum_{\pi \subseteq \Pi} \Delta_v(i, S_\pi), \tag{2}$$

where $\Pi$ is the set of all possible feature orderings (i.e., all possible feature addition sequences), and its size is $|F|!$. $S_\pi$ represents the set of features preceding feature $i$ in the ordering $\pi$. Below is the specific proof:

***Proof 1***: Let $F$ denote the feature set. For each permutation $\pi \in \Pi$, define: $S_\pi = \{j \in F\setminus\{i\} | \pi^{-1}(j) < \pi^{-1}(i)\}$. That is, $S_\pi$ contains all features that appear before $i$ in the permutation $\pi$. First, compute the number of permutations where the feature subset $S$ serves as the predecessor set of $i$. For a feature subset $S \in F\setminus\{i\}$, calculate the number of permutations where $S_\pi = S$. There are $|S|!$ ways to permute the subset $S$ while the remaining features $F\setminus(S \cup \{i\})$, can be permuted in $(|F| - |S| - 1)!$ Ways. Feature $i$ is fixed at position $|S| + 1$. Thus, the total number of permutations where $S_\pi = S$ is given by $\text{Num}(S) = |S|! \times ((|F| - |S| - 1)!)$. Next, compute the proportion of all permutations in which the feature subset $S$ occurs. Let $|F|!$ denote the total number of permutations. Then, the probability of $S$ predecessor set of $i$ is: $P(S) = \frac{\text{Num}(S)}{|F|!} = \frac{(|F|-|S|-1)!}{|F|!} = \varsigma(|S|)$. According to the definition in Exp. (1), since $P(S) = \varsigma(|S|)$, it follows that: $\sum_{S \subseteq (\mathcal{F}\setminus\{i\})} P(S) \times \Delta_v(i, S)$, note that each $\Delta_v(i, S)$ appears $\text{Num}(S)$ times across all permutations. Therefore:

$$Sv(i) = \frac{1}{|F|!} \sum_{S \subseteq (\mathcal{F}\setminus\{i\})} \text{Num}(S) \times \Delta_v(i, S) = \frac{1}{|F|!} \sum_{\pi \subseteq \Pi} \Delta_v(i, S_\pi).$$

However, this averaging method has limitations. When features are correlated, averaging Shapley Values may underestimate the contribution of important features. This occurs because, in feature permutations, if a highly correlated feature $j$ appears before $i$, the marginal contribution $\Delta_v(i, S)$ of $i$ may become small or even zero, as the correlated feature $j$ has already explained part or all the information. More specifically, two issues arise: First is Underestimation of important features. Due to feature correlation, the marginal contribution of important features is diluted during averaging. Second is incorrect feature ordering. Less significant features may receive higher Shapley Values because they lack correlated features, and their marginal contribution is not diluted during averaging. This leads to the misinterpretation illustrated in **Fig. 1** of **Section 1**, causing inaccuracies in explanations. This section will demonstrate the problems outlined in Section 1. The proof is as follows:

**Proof 2**: Let $F = \{x_1, x_2, \ldots, x_n\}$ denote the feature set. Assume feature $x_i$ is highly correlated with feature $x_j$, with correlation coefficient or similarity $s_{i,j} \approx 1$. In Shapley Value calculation, the marginal contribution $\Delta_v(i, S_\pi)$ depends on the feature set $S_\pi$ preceding $x_i$ in permutation $\pi$. When the correlated feature $x_j \notin S_\pi$ the marginal contribution of $x_i$ may be large because $x_j$ has not yet appeared, and the information of $x_i$ remains unexplained. When $x_j \in S_\pi$, due to the high correlation between $x_j$ and $x_i$, $x_j$ has already explained most of $x_i$ information. Consequently, the gain $\Delta_v(i, S_\pi)$ from adding $x_i$ may be small or even zero.

According to Exp. (2), since permutations are random, $x_j$ appears before or after $x_i$ with equal probability. Thus: when $x_j$ appears before $x_i$, the number of such permutations is approximately $\frac{|F|!}{2}$, and $\Delta_v(i, S_\pi)$ is small in these cases. When $x_j$ appears after $x_i$, the number of such permutations is approximately $\frac{|F|!}{2}$, and $\Delta_v(i, S_\pi)$ is large in these cases.

The Shapley Value can be divided into two parts: $Sv(i) = \frac{1}{2}E[\Delta_v(i, S_\pi)|x_j \notin S_\pi] + \frac{1}{2}E[\Delta_v(i, S_\pi)|x_j \in S_\pi]$. Since $\Delta_v(i, S_\pi)$ is smaller when $x_j \in S_\pi$, the overall Shapley Value $Sv(i)$ is reduced. When $x_j$ is fully correlated with $x_i$ $(S_{i,j} = 1)$, $\Delta_v(i, S_\pi) \approx 0$ when $x_j \in S_\pi$. Assume $\Delta_v(i, S_\pi) = \delta$ (a large positive value) when $x_j \notin S_\pi$, and $\Delta_v(i, S_\pi) = \epsilon$ (a value close to zero) when) when $x_j \in S_\pi$. The Shapley Value becomes: $Sv(i) = \frac{1}{2}\delta + \frac{1}{2}\epsilon \approx \frac{\delta}{2}$. This is half of the Shapley Value in the absence of correlated features ($\delta$).

## 3.2 RealExp

To address the aforementioned issues, this paper proposes a new calculation method called **RealExp**. This method decouples the Shapley Value into two parts and adjusts the calculation of marginal contributions based on feature similarity. Specifically, the Shapley Value $\emptyset_i$ of each feature is decomposed into two components: $\emptyset_i = \emptyset_i^{independent} + \emptyset_i^{Margin}$. The independent contribution $\emptyset_i^{independent}$ is represented as: $\emptyset_i^{independent} = v(\{i\}) - v(\emptyset)$, indicating the contribution of feature $x_i$ to the model when considered independently of other features. The marginal contribution $\emptyset_i^{Margin}$ is expressed as: $\emptyset_i^{Margi} = \sum_{j \neq i} w_{i,j}[v(\{i,j\}) - v(\{j\}) - \emptyset_i^{independent}]$. Where, $w_{i,j}$ reflects the similarity between features $x_i$ and $x_j$ defined as: $w_{i,j} = \frac{1-s_{i,j}}{\sum_{k \neq i}(1-s_{i,k})}$. Herein, $s_{i,j}$ represents the similarity between features $x_i$ and $x_j$, where $0 \leq s_{i,j} \leq 1$. When $s_{i,j} = 1$, indicates complete correlation between the features. When $s_{i,j} = 0$ indicates no correlation. Thus, when $s_{i,j} \approx 1$, $w_{i,j} \approx 0$, reducing the impact of correlated features on the marginal contribution. The overall definition of RealExp is given by Exp. (3):

$$\emptyset_i = \frac{1}{|F|!}\sum_{\pi \subseteq \Pi} \Delta_v(i, S_\pi) \cdot \Upsilon(S_\pi, i), \qquad (3)$$

Where $\Upsilon(\mathcal{S}_\pi, i)$ is an adjustment factor defined as: $\Upsilon(\mathcal{S}_\pi, i) = \prod_{j \in \mathcal{S}_\pi}(1 - s_{i,j})$. When $x_i$ has high similarity with a feature $x_j$ in $\mathcal{S}_\pi$, $\Upsilon(\mathcal{S}_\pi, i)$ decreases, thereby reducing the weight of $x_i$'s marginal contribution to the Shapley Value for that permutation. This adjustment also ensures that contributions from similar features are not double counted. By this design, the proposed RealExp comprehensively accounts for feature correlations and avoids the averaging issues inherent in Shapley Values.

***Proof 3:*** First, for each permutation $\pi \in \Pi$, the marginal contribution of feature $x_i$ is adjusted to $\Delta_v(i, \mathcal{S}_\pi) \cdot \Upsilon(\mathcal{S}_\pi, i)$, when $x_i$ has high similarity with a feature $x_j$ in $\mathcal{S}_\pi$ $(s_{ij} \approx 1)$: $\Upsilon(\mathcal{S}_\pi, i) = \prod_{j \in \mathcal{S}_\pi}(1 - s_{i,j}) \approx 0$, hence, the marginal contribution of $x_i$ for that permutation is significantly reduced or ignored. Next, according to Exp. (3): when $x_j \in \mathcal{S}_\pi$ and $s_{i,j} \approx 1$, $\Upsilon(\mathcal{S}_\pi, i) \approx 0$, and the corresponding marginal contribution is ignored, avoiding dilution caused by correlated features. When $x_j \notin \mathcal{S}_\pi$ and $s_{i,j}$ is small, $\Upsilon(\mathcal{S}_\pi, i)$ is relatively large, ensuring that the marginal contribution of $x_i$ is fully considered.

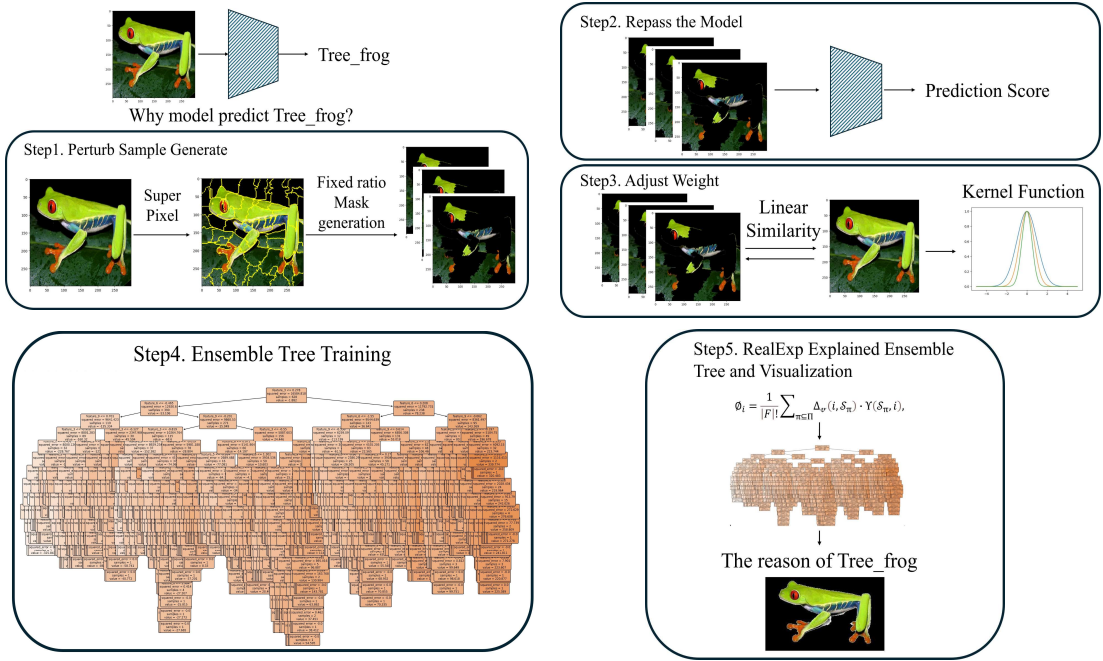

**Fig.3.** RealExp's specific workflow in practice.

### 3.3. In Practice

In practice, this paper applies RealExp to perturbation-based surrogate models to achieve interpretability, with the advantage of being able to explain any model. Unlike previous works (Ribeiro et al., 2016; Liu et al., 2024; Tan et al., 2024), which utilize random sampling and Monte Carlo sampling, this study adopts a fixed ratio to generate perturbation samples, effectively reducing instability. Furthermore, in constructing the surrogate model, this paper designs an ensemble tree as the surrogate model instead of traditional linear regression or its variants, as tree models inherently capture nonlinear relationships between features, avoiding potential collinearity issues associated with traditional linear models. Fig. 3 shows the workflow of RealExp in actual deployment. As shown in Fig. 3, taking an image classification task as an example, RealExp explains why the black-box model identifies the image as a tree frog through five steps: Step 1 involves generating perturbation samples, Step 2 re-feeds these samples into the black-box model to obtain predictions, Step 3 compares the similarity between the perturbation samples and the original image, Step 4 trains an ensemble tree model, and Step 5 uses RealExp to generate explanations and visualized outputs. The detailed process is as follows:

### 3.3.1. Perturb Sample Generation

This step is used to generate the input data for building the proxy model later. Let $\mathcal{I}$ denote the image to be interpreted and let $\mathbb{W}$ represent the watershed image segmentation algorithm. Upon segmentation, the result obtains the image $\mathcal{I}'$ such that $\mathcal{I}' = \mathbb{W}(\mathcal{I})$. The segmented image $\mathcal{I}'$ can be represented as a set of segmented blocks, denoted by $\{\mathfrak{B}_j\}_{j=1}^{n}$, where each block $\mathfrak{B}_j$ for $j \in [1, n]$ signifies the index of the block.

Let $\mathcal{f} = \{\mathfrak{y}^{(k)}\}_{k=1}^{K}$ denote a set of masks to be applied to $\mathcal{I}'$. Each mask $\mathfrak{y}^{(k)}$, where $k \in [1, K]$, contains a set of Boolean variables $\{\mu_j^{(k)}\}_{j=1}^{n}$. Each $\mu_j^{(k)}$ corresponds to block $\mathfrak{B}_j$ and indicates whether it is kept or masked (0 for masked, 1 for kept). For each specific mask application, which can be define:
$$\mathfrak{B}_j^{(k)} = \begin{cases} 0, \text{if } \mu_j^{(k)} = 0, \\ \mathfrak{B}_j, \text{if } \mu_j^{(k)} = 1. \end{cases}$$

Each set $\{\mu_j^{(k)}\}$ ensures, according to a heuristic rule, that at most 30% of the blocks are covered. This is formalized by the inequality: $\sum_{j=1}^{n}(1 - \mu_j^{(k)}) \leq 0.3n$. Applying each mask $\mathfrak{y}^{(k)}$ to $\mathcal{I}'$ yields a series of new images with selectively retained blocks: $\mathcal{I}^{(k)} = \mathfrak{y}^{(k)}(\mathcal{I}') = \cup_{j=1}^{n} \mathfrak{B}_j^{(k)}$. Collectively, these form the final set of perturbed images: $\mathcal{I}'' = \{\mathcal{I}^{(1)}, \mathcal{I}^{(2)}, \dots, \mathcal{I}^{(K)}\}$, where each image $\mathcal{I}^{(k)}$ represents the outcome of a specific block retention pattern determined by the mask $\mathfrak{y}^{(k)}$. Proof4 aims to Demonstrating the Stability of Fixed Proportion Generation.

***Proof 4 :*** Let the original image, after segmentation, be divided into $n$ blocks, denotes as $\{B_1, B_2, \dots, B_n\}$. Let $Var[f(\mathcal{I}^{(k)})]$ denote the variance of the estimated value. For fixed proportion generation, the number of masked blocks $\mathfrak{y}^{(k)}$ is fixed and denoted as $m = \alpha n$, where $\alpha \leq 0.3$. Thus, the probability of masking each block is the same and satisfies: $P(\mu_j^{(k)} = 0) = \frac{m}{n} = \alpha$, $P(\mu_j^{(k)} = 1) = 1 - \alpha$. To simplify the calculation, assume that the function $f(\mathcal{I}^{(k)})$ is linear, given by: $f(\mathcal{I}^{(k)}) = f_0 + \sum_{j=1}^{n} c_j \mu_j^{(k)}$, where $f_0$ is a constant and $c_j$ is the contribution of block $B_j$. For fixed proportion generation, the variance is given by: $Var_{fixed} = \alpha(1-\alpha)\left(\left(1 + \frac{1}{n-1}\right)\sum_{i=1}^{n} c_i^2 - \frac{1}{n-1}(\sum_{i=1}^{n} c_i)^2\right)$.

For Monte Carlo sampling, the masking probability for each block is given by $P(\mu_j^{(k)} = 0) = q$, with a variance of the masking probability $\sigma_q^2 > 0$, where $q = \alpha$. The variance is expressed as: $Var_{\text{Monte Carlo}} = (q(1-q) + \sigma_q^2)\sum_{i=1}^{n} c_i^2$. In random sampling, where each block is independently masked with probability $\alpha$ the variance is $Var_{\text{Rando}} = \alpha(1-\alpha)\sum_{i=1}^{n} c_i^2$. Comparison of Variances, it can be observed that: $Var_{fixed} < Var_{\text{Rando}} < Var_{\text{Monte Carlo}}$. This inequality demonstrates that fixed proportion generation has the lowest variance, providing greater stability compared to both random sampling and Monte Carlo sampling.

### 3.3.2. Repass the model

This step is used to generate the label for building the proxy model later. Let $\wp$ denote the model to be explained. The generated perturbed images are represented as $\mathcal{I}'' = \{\mathcal{I}_n^{(1)}, \mathcal{I}_n^{(2)}, \dots, \mathcal{I}_n^{(K)}\}$, where each $\mathcal{I}^{(k)}$ is passed through the model $\wp$ for prediction. Let $\mathcal{Y} = \{Y^{(1)}, Y^{(2)}, \dots, Y^{(K)}\}$ denote the $\wp$ for all perturbed images.

### 3.3.3. Adjust weight

This step is used to generate and adjust weight for input data in proxy model. The $\mathcal{I}'$ can be regarded as a vector of one, and the perturbed image $\mathcal{I}^{(k)}$ is represented by a combination of zeros and ones (as determined by the mask $\mu^{(k)}$, the similarity measure calculates the proportion of retained blocks. The similarity can be expressed as $sim(\mathcal{I}', \mathcal{I}^{(k)}) = \frac{1}{n}\sum_{j=1}^{n} \mu_j^{(k)} = \frac{1}{n}\|\mu^{(k)}\|_1$, where $\mu_j^{(k)}$ represents the $j$-th element in the mask $\mu^{(k)}$, and $n$ is the total number of blocks. This calculated similarity varies linearly with the number of retained blocks, which does not align with conventional nonlinear similarity measures.

To introduce nonlinearity, the similarity using an exponential function be adjusted, as shown in Exp. (4):

$$adj_k = exp\left(-\lambda\left(1 - sim(\mathcal{I}', \mathcal{I}^{(k)})\right)\right), \quad (4)$$

Where $\lambda > 0$ is a hyperparameter representing the intensity of the adjustment. The term $1 - sim(\mathcal{I}', \mathcal{I}^{(k)})$ corresponds to the normalized Hamming distance between $\mathcal{I}'$ and $\mathcal{I}^{(k)}$. This adjustment ensures that when the proportion of masked blocks is small, the adjusted similarity $adj_k$ remains high, and when the proportion is large, $adj_k$ becomes lower. The adjusted similarity $adj_k$ is applied multiplicatively to each block of the perturbed image $\mathcal{I}^{(k)}$. The updated block $B_j^{(k)\prime}$ is calculated as $B_j^{(k)\prime} = adj_k \cdot B_j^{(k)}, \forall j \in \{1,2,\dots,n\}$, where $B_j^{(k)}$ represents the block in the adjusted perturbed image, and $B_j^{(k)}$ represents the corresponding block in the original perturbed image before adjustment.

### 3.3.4. Ensemble Tree Training

This step is used to train proxy models. Till now, building upon the established perturbed sample set $\mathcal{I}'' = \{\mathcal{I}_n^{(1)}, \mathcal{I}_n^{(2)}, \dots, \mathcal{I}_n^{(K)}\}$, where each $\mathcal{I}_n^{(k)}$ represents the $k$-th perturbed sample, we associate a corresponding target value set $\mathcal{Y} = \{Y^{(1)}, Y^{(2)}, \dots, Y^{(K)}\}$. Each perturbed sample $\mathcal{I}_n^{(k)}$ comprises $n$ processed blocks $\{B'_{1,k}, B'_{2,k}, \dots, B'_{n,k}\}$, which serve as the feature set $\mathbb{x}^{(k)} = \{B'_{j,k}\}_{j=1}^{n}$.

An ensemble of regression tree models $\{\mathcal{T}_s\}_{s=1}^{S}$ is defined, where $s = 1,2,\dots,S$ and $S$ denotes the total number of trees in the ensemble.

To construct the ensemble, bootstrap sampling is performed on each feature set $\mathbb{x}^{(k)}$ to generate new training datasets $\mathcal{D}_m$. This process involves randomly drawing $N$ samples with replacement from the original $N$ samples to form each training set $\mathcal{D}_m$, which includes both input features and corresponding target outputs.

For each dataset $\mathcal{D}_m$, a regression tree $\mathcal{T}_s$ is recursively built. At each decision node within the tree, the objective is to find the optimal split by selecting the best split block $B'_{j,k}$ and the optimal split point

$L_{best}$ that minimize the weighted total variance after the split. This optimization problem is expressed as Exp. (5):

$$(B'_{j,k}, s_{best}) = \arg\min_{B'_{j,k}, s}[w_{left} \times Var(\mathcal{D}_{left}) + w_{right} \times Var(\mathcal{D}_{right})], \tag{5}$$

Where, $B'_{j,k}$ represents the j-th processed block in the k-th perturbed sample. $s$ denote a potential split point. $w_{left} = \frac{|\mathcal{D}_{left}|}{|\mathcal{D}|}$ and $w_{right} = \frac{|\mathcal{D}_{right}|}{|\mathcal{D}|}$ are the weights corresponding to the proportions of data in the left and right child nodes after the split. $Var(.)$ denotes the variance of the target values within a dataset. $\mathcal{D}_{left}$ and $\mathcal{D}_{right}$ are the subsets of $\mathcal{D}$ resulting from the split at $s$.

For each leaf node in the tree $\mathcal{T}_s$, the prediction value $\hat{y}_{leaf}$ is calculated as the meaning of all target values $y_i$ associated with the samples in that node $\hat{y}_{leaf} = \frac{1}{|\mathcal{D}_{left}|}\sum_{i \in \mathcal{D}_{leaf}} y_i$.

### 3.3.5. RealExp Explained Ensemble Tree and Visualization

This step aims to calculate the importance of the ensemble tree model using the RealExp method and output the results through visualization techniques. Specifically, each block $\mathcal{B}_i$ in the segmented set $\{\mathcal{B}_1, \mathcal{B}_2, ..., \mathcal{B}_n\}$ is treated as a feature $x_i$, and the RealExp method is applied for calculation, as described in Exp. (3). After the computation, the importance scores are represented as $\{\emptyset_1, \emptyset_2, ..., \emptyset_n\}$. These scores are then sorted in descending order, resulting in $\emptyset_{i1} \geq \emptyset_{i2} \geq \cdots \geq \emptyset_{in}$, where $\emptyset_{ik}$ denotes the Shapley Value of the k-th block in the sorted sequence. Let $M: \mathbb{R} \to \mathcal{C}$ denote a mapping function which maps numerical importance scores to visual attributes, thereby intuitively presenting the importance of each block.

## 4. Assumptions and Problem Formulation

This section introduces the assumptions and problem statements of this paper. The evaluation incorporates Kendall's $\tau$ coefficient to further analyze the feature importance paths, combining the calculation of feature importance paths and accuracy to provide deeper insights into the model's decision-making process. This design aims to help users understand whether the model's reasoning aligns with the fundamental understanding of experts and to assess the model's consistency and accuracy in feature selection and classification decisions.

For image classification tasks, let $\mathcal{M}$ denote the model to be explained, and the input image be $\mathcal{I} \in \mathbb{R}^{H \times W \times C}$, where $H$、$W$ and $C$ represent the height, width, and channel number of the image, respectively. Let $\mathcal{Q}$ denote the image segmentation algorithm, which divides $\mathcal{I}$ into $n$ superpixel regions, resulting in the segmented set: $\mathcal{I} = \bigcup_{i=1}^{n} S_i$. Where $S_i$ represent the $i$-th region, satisfying $S_i \cap S_j = \phi$ when $i \neq j$.

Let $\mathcal{U} = \{S_{u1}, S_{u2}, ..., S_{um}\}$ denote the set of superpixels identified by experts as significantly contributing to the classification result, where $\mathcal{U} \subseteq \{S_1, S_2, ..., S_n\}$, and $m \leq n$. Through an explainable algorithm, the model $\mathcal{M}$ assigns an importance score $f(S_i)$ to each superpixel $S_i$。Let $\mathcal{K} = \{S_{k1}, S_{k2}, ..., S_{km}\}$ represent the superpixels sorted in descending order of importance, where $\mathcal{K} \subseteq$

$\{S_1, S_2, \ldots, S_n\}$. To quantify the consistency between the model's and the expert's feature importance rankings, here are Consistency Evaluation Steps: First is Matching Count Calculation: Compare the overlap between $\mathcal{U}$ and $\mathcal{K}$, denote as in Exp. (6):

$$\mathcal{H} = \mathcal{U} \cap \mathcal{K}. \tag{6}$$

The matching count $|\mathcal{H}|$ represents how many of the expert-selected blocks are correctly identified by the model. Second is Ranking Consistency Calculation: For all blocks in $\mathcal{H}$, calculate Kendall's $\tau$ coefficient to evaluate ranking consistency in Exp. (7):

$$\tau = \frac{2(C-D)}{|\mathcal{H}|(|\mathcal{H}|-1)}, \tag{7}$$

Where $C$ represent the number of concordant pairs, where for any two blocks $S_{hi}$ and $S_{hj}$ in $\mathcal{H}$ satisfy $i < j$, and their relative order in $\mathcal{K}$ also satisfies $\mathcal{K}_i < \mathcal{K}_j$. $D$ represent the number of discordant pairs, where any two blocks $S_{hi}$ and $S_{hj}$ in $\mathcal{U}$ satisfy $i < j$, their relative order in $\mathcal{K}$ satisfies $\mathcal{K}_i > \mathcal{K}_j$. Let $M_{best}$ denote the optimization objective, aiming to find the optimal explanation mechanism $M_{best}$ that maximizes $\tau$, as defined by $M_{best} = \arg\max_M \tau(\mathcal{U}, \mathcal{K})$.

Suppose the image is segmented into $n = 10$ regions $S_1, S_2, \ldots, S_{10}$. The 5 most important blocks selected by the experts are $\mathcal{U} = \{S_1, S_3, S_5, S_7, S_9\}$, ranked as $S_1 > S_3 > S_5 > S_7 > S_9$. The model identifies the top 5 blocks as $\mathcal{K} = \{S_3, S_1, S_7, S_5, S_2\}$.

For Matching Count Calculation using Exp. (6), the matching count $|\mathcal{H}| = 4$, indicating the model correctly identified 4 of the 5 expert-selected blocks. The accuracy is 0.8.

For Ranking Consistency Calculation using Exp. (7), the Kendall's $\tau$ coefficient is computed as $\tau = \frac{2(C-D)}{|\mathcal{H}|(|\mathcal{H}|-1)} = \frac{2(4-2)}{4(4-1)} = \frac{4}{12} = 0.333$.

For text sentiment analysis, given a model $\mathfrak{M}$ that needs to be explained and an input text $\mathfrak{T}$, the output is the set of words in the text on which the model bases its decisions. Taking Chinese text as an example, let $\mathfrak{S}$ denote the Jieba segmentation algorithm, and let the segmented sentence be represented as $\mathfrak{T} = \{T_1, T_2, \ldots, T_n\}$, where $T_i$ denotes the $i$-th word or token after segmentation, and $n$ is the total number of words in the text.

Apply masking perturbation techniques to $\mathfrak{T}$ to generate a set of perturbed texts. Let the set of perturbed samples be denoted as $\mathfrak{T}' = \{\mathfrak{T}'^{(1)}, \mathfrak{T}'^{(2)}, \ldots, \mathfrak{T}'^{(K)}\}$, where $K$ represents the number of masking perturbations applied. Each perturbed sample $\mathfrak{T}'^{(k)}$ is a subset of the original text $\mathfrak{T}$, constructed by masking certain words. Specifically, each $\mathfrak{T}'^{(k)}$ is defined as $\mathfrak{T}'^{(k)} = \{T_{j_1}^{(k)}, T_{j_2}^{(k)}, \ldots, T_{j_p}^{(k)}\}$, with $0.3n < p \leq n$, where $\{j_1^{(k)}, j_2^{(k)}, \ldots, j_p^{(k)}\} \subseteq \{1, 2, \ldots, n\}$ indicates the indices of the unmasked words in the $k$-th perturbation.

For each perturbed text $\mathfrak{T}'^{(k)}$, the model $\mathfrak{M}$ is used to obtain the prediction score $Y^{(k)}$. The set of prediction scores is represented as $\mathcal{Y} = \{Y^{(1)}, Y^{(2)}, \ldots, Y^{(K)}\}$. The similarity between each perturbed sample $\mathfrak{T}'^{(k)}$ and the original text $\mathfrak{T}$ is calculated using a similarity function $sim(\mathfrak{T}, \mathfrak{T}'^{(k)}) = \frac{1}{n}|\mathfrak{T}'^{(k)}| = \frac{p}{n}$, where $|\mathfrak{T}'^{(k)}| = p$ is the number of unmasked words in $\mathfrak{T}'^{(k)}$. To adjust for the linearity of the similarity measure, an exponential adjustment is applied in $w^{(k)} = exp\left(-\lambda\left(1-\right.\right.$

$sim(\mathfrak{T}, \mathfrak{T}'^{(k)})))$.

Construct the input feature set $\mathcal{X}$ for the proxy model by weighting the perturbed samples $\mathcal{X} = \{\mathfrak{T}'^{(1)} \times w^{(1)}, \mathfrak{T}'^{(2)} \times w^{(2)}, ..., \mathfrak{T}'^{(K)} \times w^{(K)}\}$, where $\mathfrak{T}'^{(K)} \times w^{(K)}$ denotes that each word in $\mathfrak{T}'^{(K)}$ is associated with the weight $w^{(k)}$. The corresponding target labels are the prediction scores obtained from the model $\mathfrak{M}$: $\mathcal{y} = \{Y^{(1)}, Y^{(2)}, ..., Y^{(K)}\}$. By training the interpretable model on the weighted input-output pairs ($\mathcal{X}$, $\mathcal{y}$), and using Exp.(3) calculated, the importance scores for each word $T_i$ in the original text $\mathfrak{T}$ can be derived. Let the set of importance scores be denoted as $P = \{\emptyset_1, \emptyset_2, ..., \emptyset_n\}$, where $\emptyset_i$ represents the importance of word $T_i$ as determined by the model. Let $\mathcal{U} = \{T_{u1}, T_{u2}, ..., T_{um}\}$ denote the set of words selected by experts as being most indicative of the classification, where $\mathcal{U} \subseteq \{T_1, T_2, ..., T_n\}$, and $m \leq n$.

Based on the computed importance scores $P$, the top $m$ words are selected and ordered to form the set $\mathcal{K} = \{T_{k1}, T_{k2}, ..., T_{km}\}$, where the words are sorted such that $\emptyset_{k1} \geq \emptyset_{k2} \geq \cdots \geq \emptyset_{km}$. Calculating Matching Score and Kendall's $\tau$ Coefficient by Exp. (6) and Exp. (7). Here is an example: Consider the input text $\mathfrak{T}$ = "這部電影很有趣，演員的表演令人印象深刻", which translates to "This movie is very interesting, and the performances by the actors are impressive." The model to be explained is a Support Vector Machine (SVM) classifier that predicts this sentence as expressing a positive sentiment. Using the Jieba segmentation algorithm $\mathfrak{S}$, the sentence is segmented into:

$$\mathfrak{T} = \{T_1 = \text{"這部"}, T_2 = \text{"電影"}, T_3 = \text{"很"}, T_4 = \text{"有趣"}, T_5 = \text{"演員"}, T_6 = \text{"的"}, T_7 = \text{"表演"}, T_8 = \text{"令人"}, T_9 = \text{"印象"}, T_{10} = \text{"深刻"}\}.$$

Let the number of perturbed samples be $K = 3$. The perturbed texts are:

$$\mathfrak{T}'^{(1)}: \{T_1, T_2, T_3\}, \{T_5, T_6, T_7, T_8, T_9, T_{10}\}, \mathfrak{T}'^{(2)}: \{T_1, T_2, T_3, T_4\}, \{T_6, T_7, T_8, T_9\}, \mathfrak{T}'^{(3)}: \{T_1, T_2, T_3, T_4\}, \{T_5, T_6, T_7, T_8, T_9, T_{10}\}$$

The prediction scores obtained from the SVM are $\mathcal{y} = \{Y^{(1)} = 0.3, Y^{(2)} = 0.7, ..., Y^{(3)} = 0.8\}$. The similarities between each perturbed sample and the original text are $sim(\mathfrak{T}, \mathfrak{T}'^{(1)}) = 0.9$, $sim(\mathfrak{T}, \mathfrak{T}'^{(2)}) = 0.8$ and $sim(\mathfrak{T}, \mathfrak{T}'^{(1)}) = 1$, Applying the exponential adjustment, the adjusted weights are $w^{(1)} = \exp(-\lambda(1 - 0.9))$, $w^{(2)} = \exp(-\lambda(1 - 0.8))$, $w^{(3)} = \exp(-\lambda(1 - 1)) = 1$.

The inputs to the proxy model are $\mathcal{X} = \{\mathfrak{T}'^{(1)} \times w^{(1)}, \mathfrak{T}'^{(2)} \times w^{(2)}, \mathfrak{T}'^{(3)} \times w^{(3)}\}$ with $\mathcal{y}$. After training, RealExp calculated the proxy model's the importance scores $P = \{\emptyset_1, \emptyset_2, ..., \emptyset_{10}\}$ for each word $T_i$ in $\mathfrak{T}$ are obtained. Experts consider the positive sentiment words in the sentence to be $\mathcal{U} = \{T_4, T_{10}\}$, and they rank them as: $T_4 > T_{10}$. Based on the importance scores $P$, the model's top two words are $\mathcal{K} = \{T_4, T_{10}\}$, with importance scores $\emptyset_{10} > \emptyset_4$. The matching words are $\mathcal{M} = 2$. The accuracy is 1. The Kendall's τ Coefficient is 1.

## 5. Experiment

This section will introduce the performance of the proposed RealExp interpretable method. All the information for the experiment is on the following the URL[1]. The experiment is divided into two parts:

---

[1] https://github.com/1846659840/The-Research-on-a-new-LIME-based-interpretable-machine-learning-and-its-evaluation-

image classification and text sentiment analysis.

### 5.1. Image Classification

This section introduces the explainable experiments of image classification, which are divided into three parts: dataset and evaluation metrics, parameter settings, and model evaluation.

### 5.5.1. Dataset and Evaluation Metrics

This paper selects two datasets for study: Animal-10[2] and Animal-151[3]. The Animal-10 dataset contains 10 different types of animals, while the Animal-151 dataset includes 15 mainstream animal categories, each comprising 10 different rare species under each category. For both datasets, 1,000 images were selected, resulting in a total of 2,000 images for annotation and model testing.

In terms of model selection, this study chose five commonly used pre-trained models: MobileNet (Andrew et al., 2017), ResNet (He et al., 2016), DenseNet (Huang et al., 2017), and Vision Transformer (ViT) (Dosovitskiy et al., 2021).

For evaluation metrics, the study focuses on four aspects: the stability of explanations, the interpretability of the model, consistency with expert understanding, and the interpretability of the decision-making path.

To ensure the stability of explanations, and consistent with previous works (Tan et al., 2024; Li et al., 2023; David et al., 2024; Liu et al., 2024), this study uses the Jaccard coefficient as the evaluation metric. The specific formula is $J = \frac{|A \cap B|}{|A \cup B|}$, where $A$ and $B$ represent the sets of results from two model outputs.

To ensure the interpretability of the model, and consistent with previous works (Tan et al., 2024; Li et al., 2023; David et al., 2024; Liu et al., 2024), this study uses the $R^2$ coefficient as the evaluation metric. The specific formula is $R^2 = 1 - \frac{\sum_{i=1}^{n}(y_i - \widehat{y_i})^2}{\sum_{i=1}^{n}(y_i - \hat{y})^2}$, where $y_i$ denotes the actual value, $\widehat{y_i}$ denotes the predicted value, and $\hat{y}$ denotes the mean of the actual values.

For the consistency with expert understanding, this study employs the $\mathcal{H}$-score described in Section 4 to calculate accuracy, as detailed in Exp. (6). Additionally, for consistency with expert understanding, Kendall's τ coefficient is used to calculate the $M$-score, as detailed in Exp. (7).

### 5.5.2. Parameter settings

This section introduces the hyperparameter settings. First, for the perturbation samples $\mathcal{J}'' = \{\mathcal{J}^{(1)}, \mathcal{J}^{(2)}, \dots, \mathcal{J}^{(K)}\}$, the number $K$ is fixed at 500, and the hyperparameter $\lambda$ in Exp. (4) is set to 0.25. This is consistent with previous works (Tan et al., 2024; Li et al., 2023; David et al., 2024; Liu et al., 2024). Second, for the pre-trained model weights, this study uniformly uses weights trained on ImageNet.

---

method-imagepart.

[2] https://www.kaggle.com/datasets/alessiocorrado99/animals10

[3] https://www.kaggle.com/datasets/sharansmenon/animals141

For the construction of the tree model, the number of $\{\mathcal{T}_s\}_{s=1}^{S}$ is also fixed at 50.

**Table 2.**

Compare the state of arts on Animal-10 Dataset

| Animal-10 | Zeiler & Fergus, 2014 (Baseline) | | | | Riberiro et al.,2016 (LIME) | | | |
|---|---|---|---|---|---|---|---|---|
| | $J$ | $R^2$ | $\mathcal{H}$-score | $M$-score | $J$ | $R^2$ | $\mathcal{H}$-score | $M$-score |
| MobileNet | 0.46 | 0.53 | 0.54 | 0.50 | 0.73 | 0.76 | 0.68 | 0.65 |
| ResNet | 0.46 | 0.51 | 0.45 | 0.50 | 0.73 | 0.77 | 0.71 | 0.60 |
| DenseNet | 0.46 | 0.52 | 0.50 | 0.50 | 0.73 | 0.76 | 0.76 | 0.58 |
| ViT | 0.46 | 0.50 | 0.50 | 0.5 | 0.73 | 0.83 | 0.81 | 0.52 |
| | Lundberg & Lee, 2017 (Kernel-SHAP) | | | | Li et al., 2023 ($\mathcal{G}$LIME) | | | |
| | $J$ | $R^2$ | $\mathcal{H}$-score | $M$-score | $J$ | $R^2$ | $\mathcal{H}$-score | $M$-score |
| MobileNet | 0.71 | 0.77 | 0.69 | 0.66 | 0.74 | 0.77 | 0.69 | 0.68 |
| ResNet | 0.71 | 0.78 | 0.72 | 0.61 | 0.74 | 0.78 | 0.72 | 0.63 |
| DenseNet | 0.71 | 0.77 | 0.76 | 0.59 | 0.74 | 0.77 | 0.76 | 0.61 |
| ViT | 0.71 | 0.84 | 0.82 | 0.53 | 0.74 | 0.84 | 0.82 | 0.55 |
| | Liu et al., 2024 (HIDTWM) | | | | David et al., 2024 (AcME) | | | |
| | $J$ | $R^2$ | $\mathcal{H}$-score | $M$-score | $J$ | $R^2$ | $\mathcal{H}$-score | $M$-score |
| MobileNet | 0.75 | 0.78 | 0.70 | 0.67 | 0.78 | 0.82 | 0.75 | 0.69 |
| ResNet | 0.75 | 0.79 | 0.73 | 0.62 | 0.78 | 0.83 | 0.78 | 0.64 |
| DenseNet | 0.75 | 0.78 | 0.78 | 0.60 | 0.78 | 0.82 | 0.83 | 0.62 |
| ViT | 0.75 | 0.85 | 0.83 | 0.54 | 0.78 | 0.90 | 0.88 | 0.56 |
| | Tan et al., 2024 (GLIME) | | | | RealEXP | | | |
| | $J$ | $R^2$ | $\mathcal{H}$-score | $M$-score | $J$ | $R^2$ | $\mathcal{H}$-score | $M$-score |
| MobileNet | 0.77 | 0.80 | 0.73 | 0.68 | 0.83 | 0.85 | 0.78 | 0.71 |
| ResNet | 0.77 | 0.81 | 0.76 | 0.63 | 0.83 | 0.86 | 0.81 | 0.66 |
| DenseNet | 0.77 | 0.80 | 0.81 | 0.61 | 0.83 | 0.85 | 0.86 | 0.64 |
| ViT | 0.77 | 0.88 | 0.86 | 0.55 | 0.83 | 0.94 | 0.92 | 0.58 |

### 5.5.3. Model Evaluation

This section introduces the Model Evaluation in Animal-10 and Animal-151. Tables 2 and 3 present a comparison of the proposed RealEXP method with other state-of-the-art approaches (including Zeiler & Fergus, 2014, LIME (Ribeiro et al., 2016), Kernel-SHAP (Lundberg & Lee, 2017), GLIME (Li et al., 2023), HIDTWM (Liu et al., 2024), AcME (David et al., 2024), and G-LIME (Tan et al., 2024)) on the Animal-10 and Animal-151 benchmark datasets. Overall, the proposed RealEXP demonstrates superior stability, enhanced interpretability of models, and better alignment with expert judgments across both datasets.

Specifically, in terms of stability (measured by the stability index $J$), RealEXP stands out due to its novel perturbation sampling strategy. Unlike traditional methods that rely on random perturbations or Monte Carlo-based sampling, RealEXP employs a fixed-proportion perturbation approach, which significantly reduces the variance of the perturbed samples. This ensures consistent explanation results. Experimental results on stability also validate the theoretical conclusion presented in Proof 4, namely, $Var_{fixed} < Var_{\text{Rand}} < Var_{\text{Monte Carlo}}$.

**Table 3.**

Compare the state of arts on Animal-151 Dataset

| Animal-151 | Zeiler & Fergus, 2014 (Baseline) | | | | Riberiro et al., 2016 (LIME) | | | |
|---|---|---|---|---|---|---|---|---|
| | $J$ | $R^2$ | $\mathcal{H}$-score | $M$-score | $J$ | $R^2$ | $\mathcal{H}$-score | $M$-score |
| MobileNet | 0.46 | 0.43 | 0.42 | 0.42 | 0.73 | 0.64 | 0.53 | 0.55 |
| ResNet | 0.46 | 0.44 | 0.30 | 0.41 | 0.73 | 0.68 | 0.53 | 0.53 |
| DenseNet | 0.46 | 0.41 | 0.36 | 0.45 | 0.73 | 0.62 | 0.71 | 0.36 |
| ViT | 0.46 | 0.44 | 0.30 | 0.45 | 0.73 | 0.67 | 0.61 | 0.25 |
| | Lundberg & Lee, 2017 (Kernel-SHAP) | | | | Li et al., 2023 ($\mathcal{G}$LIME) | | | |
| | $J$ | $R^2$ | $\mathcal{H}$-score | $M$-score | $J$ | $R^2$ | $\mathcal{H}$-score | $M$-score |
| MobileNet | 0.71 | 0.63 | 0.51 | 0.54 | 0.74 | 0.67 | 0.51 | 0.54 |
| ResNet | 0.71 | 0.69 | 0.52 | 0.54 | 0.74 | 0.66 | 0.52 | 0.55 |
| DenseNet | 0.71 | 0.62 | 0.66 | 0.37 | 0.74 | 0.62 | 0.71 | 0.39 |
| ViT | 0.71 | 0.68 | 0.63 | 0.42 | 0.74 | 0.68 | 0.63 | 0.44 |
| | Liu et al., 2024 (HIDTWM) | | | | David et al., 2024 (AcME) | | | |
| | $J$ | $R^2$ | $\mathcal{H}$-score | $M$-score | $J$ | $R^2$ | $\mathcal{H}$-score | $M$-score |
| MobileNet | 0.75 | 0.66 | 0.55 | 0.57 | 0.78 | 0.62 | 0.65 | 0.54 |
| ResNet | 0.75 | 0.71 | 0.55 | 0.53 | 0.78 | 0.69 | 0.60 | 0.55 |
| DenseNet | 0.75 | 0.58 | 0.73 | 0.38 | 0.78 | 0.57 | 0.73 | 0.57 |
| ViT | 0.75 | 0.69 | 0.64 | 0.43 | 0.78 | 0.71 | 0.66 | 0.45 |
| | Tan et al., 2024 (GLIME) | | | | RealEXP | | | |
| | $J$ | $R^2$ | $\mathcal{H}$-score | $M$-score | $J$ | $R^2$ | $\mathcal{H}$-score | $M$-score |
| MobileNet | 0.77 | 0.61 | 0.59 | 0.43 | **0.83** | 0.73 | 0.56 | **0.51** |
| ResNet | 0.77 | 0.65 | 0.54 | 0.50 | **0.83** | 0.66 | 0.65 | 0.51 |
| DenseNet | 0.77 | 0.69 | 0.54 | 0.46 | **0.83** | 0.70 | 0.67 | 0.52 |
| ViT | 0.77 | 0.65 | 0.66 | 0.41 | **0.83** | **0.84** | **0.82** | 0.48 |

For model interpretability (measured by $R^2$), the proposed RealEXP again demonstrates significant improvements. This is primarily attributed to its innovative decomposition of Shapley Value into independent contributions and interaction contributions. Compared to methods like Kernel-SHAP and AcME, which rely on the traditional Shapley Value for interpretability, RealEXP places greater emphasis on feature importance, leading to more accurate and insightful explanations of deep learning models.

Finally, RealEXP exhibits enhanced performance in aligning with expert evaluations, as reflected by higher $\mathcal{H}$-score and $M$-score. This improvement is due to its thoughtful selection of proxy models and feature importance evaluation. Unlike GLIME, which directly outputs slopes using Gaussian ridge regression, RealEXP incorporates an ensemble tree model to capture nonlinear relationships between features. After training the tree model, RealEXP further assesses the feature importance within the ensemble, achieving more robust interpretability and better alignment with the underlying model behavior.

In the evaluation of the interpretability of four pre-trained models, an intriguing phenomenon was observed: ViT outperformed convolutional networks in $\mathcal{H}$-score, demonstrating its advantage in overall task performance. However, in terms of $M$-score, convolutional networks achieved higher scores, indicating that their decision-making paths are more aligned with human expert reasoning. This phenomenon reflects the differences in the design philosophies of these two models: ViT emphasizes global representation, leveraging multi-head attention mechanisms to model relationships between patches, enabling it to capture global semantic information. This contributes to its superior overall performance. However, the global mechanism of ViT often distributes attention in a way that deviates from human intuition, leading to weaker interpretability. In contrast, convolutional networks, by sequentially capturing local features, align more closely with human observation patterns, thereby exhibiting better interpretability in decision-making paths. Nevertheless, due to the cumulative nature of local feature extraction, convolutional networks struggle to integrate global context effectively, resulting in lower overall performance compared to ViT.

Furthermore, the choice of model should also depend on the specific task context. For example, in video classification tasks requiring frame-level feature extraction, ViT's ability to model global relationships makes it more effective at capturing temporal consistency. On the other hand, for traditional image classification tasks, especially those involving fine-tuning of pre-trained models, convolutional networks often excel due to their ability to capture richer fine-grained details. These observations provide valuable insights for future model design. For instance, combining the strengths of ViT and convolutional networks and employing multi-scale feature representation methods could achieve a balance between global semantic understanding and local feature extraction, further enhancing the overall performance and interpretability of models.

Figure 4 visually presents the results of applying different explainability methods on the ViT pre-trained model. It is evident that the proposed RealEXP method demonstrates higher stability and model interpretability compared to other explainability approaches. This is primarily due to two key factors: First, RealEXP places greater emphasis on feature importance, leading to more accurate and insightful explanations of deep learning models. Second, the RealEXP method prioritizes dynamic weight allocation and regional sensitivity optimization when constructing explanations. Furthermore, in multiple perturbation experiments, the RealEXP method consistently identifies regions closely related to the target classification, whereas other methods, such as LIME, Kernel-SHAP, and GLIME, may exhibit feature region drift or inconsistency under certain conditions. This indicates that RealEXP not only provides more reliable model explanations but also better meets the demand for model transparency in practical applications.

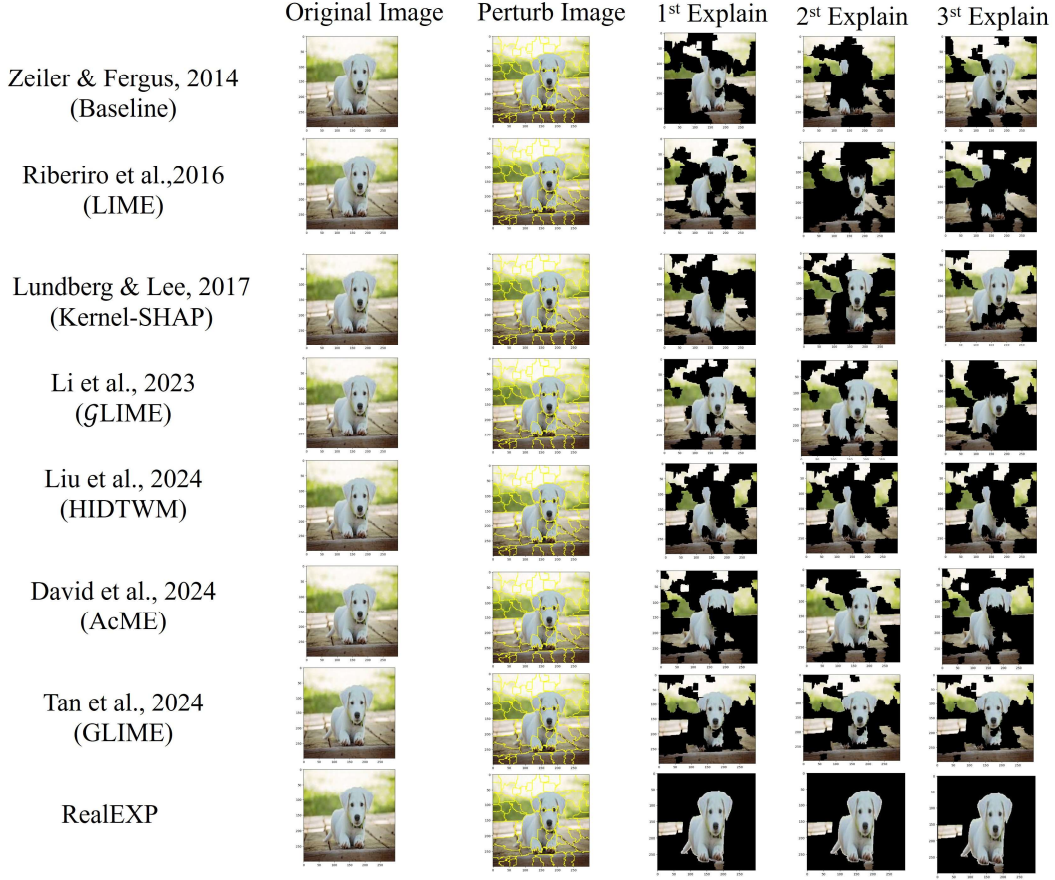

**Fig.4.** Example of Explainable Visualization for ViT

**Table 3.**

Ablation Study in Animal-151 Dataset in ViT

| Fixed Perturb | Ensemble Tree | RealEXP | $J$ | $R^2$ | $\mathcal{H}$-score | $M$-score |
|---|---|---|---|---|---|---|
| ✔ | | | 0.73 | 0.63 | 0.30 | 0.33 |
| | ✔ | | 0.46 | 0.67 | 0.64 | 0.32 |
| | | ✔ | 0.42 | 0.68 | 0.64 | 0.39 |
| ✔ | ✔ | | 0.73 | 0.68 | 0.61 | 0.27 |
| ✔ | | ✔ | 0.71 | 0.72 | 0.68 | 0.42 |
| | ✔ | ✔ | 0.48 | 0.54 | 0.67 | 0.33 |
| ✔ | ✔ | ✔ | **0.83** | **0.84** | **0.82** | **0.48** |

Table 3 presents the ablation study results on the Animal-151 dataset based on the ViT model, evaluating the impact of different method combinations on model interpretability. Among single methods, Fixed Perturb showed outstanding performance in the Jaccard coefficient $J$ (0.73), while Ensemble Tree and RealEXP contributed moderately to $\mathcal{H}$-score (0.64) and $M$-score (0.39), respectively, but their overall effectiveness was relatively limited. For dual-method combinations, the pairing of Fixed Perturb and RealEXP demonstrated the best results, significantly improving $R^2$ (0.72) and $M$-score (0.42). However, when all three methods (Fixed Perturb, Ensemble Tree, and RealEXP) were used together, all metrics reached their highest values: J at 0.83, $R^2$ at 0.84, $\mathcal{H}$-score at 0.82, and $M$-score at 0.48. This indicates that the combination of all three methods maximizes interpretability, stability, and overall performance, highlighting the importance of their synergistic effects.

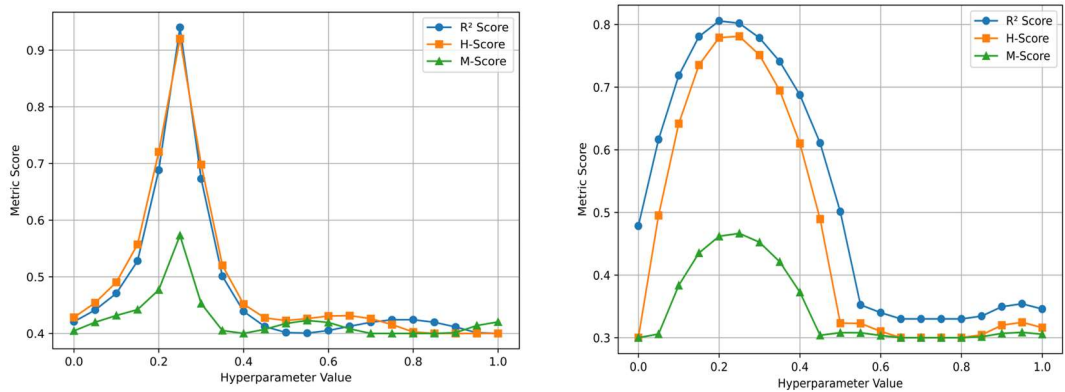

(a). Animal-10                    (b). Animal-151

**Fig.5.** hyperparameter $\lambda$ setting in Exp. (4).

Figure 5 aims to discuss the performance of RealEXP under different hyperparameter $\lambda$ settings. It can be observed that variations in $\lambda$ significantly impact the model's performance, with the best performance achieved when $\lambda = 0.25$. This is because, at $\lambda = 0.25$, a good balance is struck between regularization strength and the model's learning capacity, effectively preventing overfitting while ensuring sufficient learning of data features.

Figure 6 aims to illustrate the model's performance under different ratio settings. The best performance is achieved when the Fixed Ratio equals 0.3. This is because, at this ratio, the distribution of data samples aligns most closely with the target task, allowing the model to learn key features more effectively while avoiding issues such as information loss or noise amplification caused by excessively high or low ratios. The above is the experimental part of images, followed by the experiments of text Sentiment Analysis.

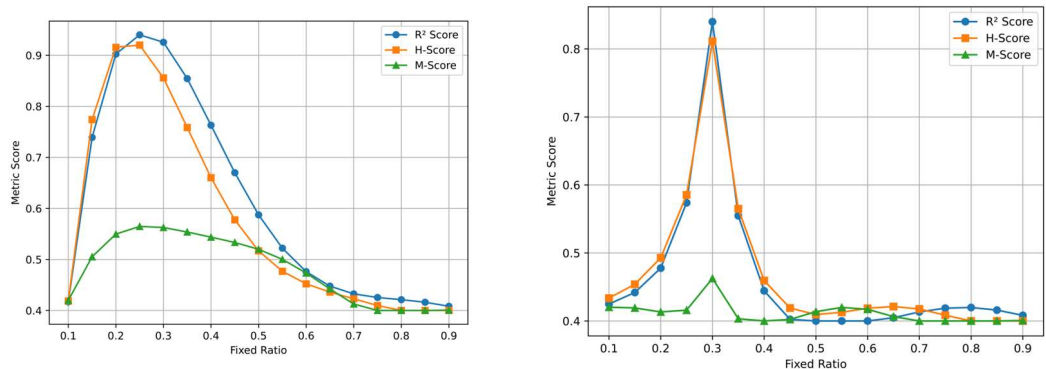

(a). Animal-10                    (b). Animal-151

**Fig.6.** Different Fixed Ratio.

## 5.2. Text Sentiment Analysis

This section introduces the explainable experiments of text sentiment analysis. The aim of this experiment is to address problems in real-life scenarios, such as detecting bullying in school settings (e.g., identifying abusive language) or filtering sensitive words in private online forums. These scenarios typically require the deployment of edge computing and localized models while demanding fast system response times. Although current Transformer-based large language models (LLMs) have achieved significant success in text processing, their large parameter sizes and reliance on extensive training data make their effective deployment in the above scenarios challenging.

Therefore, improvements in interpretable machine learning that enable traditional Bag-of-Words (BoW) methods to deliver good performance while remaining lightweight would greatly facilitate the successful deployment and application of these models in such scenarios. Specifically, if traditional BoW models can achieve performance comparable to LLMs in detecting sensitive words and specific terms, their success rate in practical applications would be further enhanced.

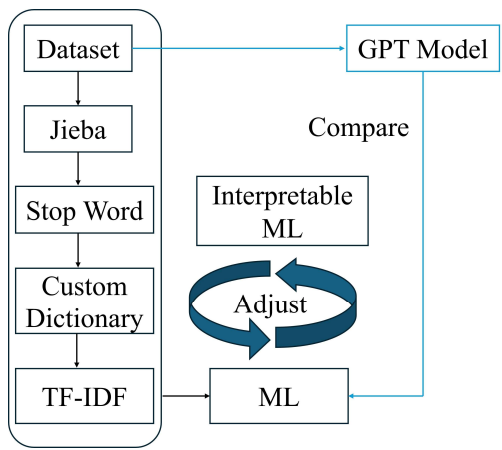

(a). Experimental workflow    (b). GPT-finetuning Model Series

**Fig.7.** Text Sentiment Analysis Experimental

Figure 7 illustrates the workflow of the text analysis experiment. Specifically, subfigure (a) explains the detailed steps of the experiment: the dataset is first segmented using Jieba, followed by further cleansing using a stop-word list and a custom dictionary. Next, word vector transformation is performed using the TF-IDF method, and the transformed data is used to train a support machine learning (ML) model. After the model is trained, various interpretable machine learning algorithms are applied to conduct interpretative analysis of the ML model, comparing the interpretable performance of each algorithm. Based on the conclusions drawn from the best interpretative algorithm, the dataset and model are reviewed and corrected, and the revised results are compared with those of the GPT model.

The dataset used in the experiment is ChnSentiCorp_htl_all[4], from which the first 1,000 positive reviews and 1,000 negative reviews were selected, totaling 2,000 sentences. For interpretability, these sentences were annotated using a bag-of-words model to identify words deemed positive or negative in each sentence. The comparison model is a fine-tuned version of the GPT series, with subfigure (b) showcasing different versions of the GPT model.

Table 4 aims to demonstrate the performance of different machine learning models on the ChnSenti-
**Table 4.**

---

[4] https://github.com/InsaneLife/ChineseNLPCorpus

Different machine Learning method on the ChnSentiCorp_htl_all

| Random Forest | Precision | Recall | F1-Score | Accuracy |
|---|---|---|---|---|
| 0 (Negative) | 0.84 | 0.90 | 0.87 | 0.87 |
| 1 (Positive) | 0.89 | 0.91 | 0.90 | |
| Logistic Regression | Precision | Recall | F1-Score | Accuracy |
| 0 (Negative) | 0.85 | 0.90 | 0.87 | 0.87 |
| 1 (Positive) | 0.89 | 0.84 | 0.86 | |
| Decision Tree | Precision | Recall | F1-Score | Accuracy |
| 0 (Negative) | 0.81 | 0.82 | 0.81 | 0.81 |
| 1 (Positive) | 0.82 | 0.81 | 0.82 | |

**Table 5.**

Different Interpretable Machine learning algorithm performance $\mathcal{H}$-score

| $\mathcal{H}$-score | LIME | Kernel-SHAP | GLIME | RealEXP |
|---|---|---|---|---|
| Random Forest | 0.3264 | 0.3549 | 0.4083 | 0.4352 |
| Logistic Regression | 0.7184 | 0.7564 | 0.8019 | 0.8198 |
| Decision Tree | 0.2736 | 0.3289 | 0.3689 | 0.3879 |

Corp_htl_all datasets. As shown in the table, Random Forest and Logistic Regression achieve the best performance, followed by Decision Tree. Table 5 aims to compare the proposed RealEXP with related methods, including LIME (Riberiro et al., 2016), Kernel-SHAP (Lundberg & Lee, 2017), and G-LIME (Tan et al., 2024), in terms of their performance on $\mathcal{H}$-Score. As shown in the figure, the proposed RealEXP significantly outperforms other interpretability algorithms in $\mathcal{H}$-Score. This is because RealEXP combines the advantages of both global and local interpretability by introducing a multi-level feature weight analysis mechanism, allowing it to capture key features in text tasks more precisely while achieving superior interpretative quality compared to traditional methods.

Additionally, an interesting phenomenon was observed: why do Decision Trees, which are considered highly interpretable, and Random Forests, which possess some interpretive capabilities, perform so poorly in interpretability scores for NLP tasks? This is because, after processing perturbed data through the tree models, some perturbed samples produced prediction results consistent with the original samples. The reason lies in the token count, which increases significantly after segmentation in Chinese text. When constructing tree models, many tokens are ignored due to their low attribute importance. Alternatively, the test data may include tokens that were not present in the training data. This causes tree models to rely on the distribution of the parent data rather than effectively explaining the input data based on specific feature weights.

To address this issue, a two-stage model, namely the RL model, which combines the Random Forest and Logistic Regression models, is proposed. Specifically, in Random Forest, if perturbed samples are found to have the same prediction probabilities as the original samples, those data points are retained. After traversing all the data, the data points with identical prediction probabilities are used to train the

Logistic Regression model.

For the trained Random Forest model, a total of 2k data points were examined, among which 285 instances were found where perturbed samples shared the same prediction probabilities as the original samples. Subsequently, a Random Forest model with 1715 data points was retrained, and the outputs of the new Random Forest were used to retrain the Logistic Regression model. During this process, the issue did not reoccur.

Additionally, the RL model achieved significant improvements in both model performance and interpretability. Specifically, the interpretability $\mathcal{H}$-score improved from 0.4352 to 0.9246 using RealEXP, while the F1 score of the combined model reached 0.92 in performance evaluation. Table 6 presents the performance of various fine-tuned models of GPT-3. As shown in Table 6, the performance of the redesigned combined machine learning model, which leverages interpretable machine learning, surpasses that of the pre-trained ada model. This indicates that by incorporating interpretability mechanisms, it is possible to optimize the model's structure and feature selection without relying on large-scale pre-training, thereby enhancing the model's performance and generalization ability. Furthermore, this demonstrates that in specific tasks, traditional machine learning methods, when combined with interpretability analysis, can effectively bridge the gap with large-scale pre-trained models and, in some cases, even achieve superior performance.

**Table 6**

Compared with GPT-3 finetuning model

| Davinci | Precision | Recall | F1 | Accuracy | Cost |
|---|---|---|---|---|---|
| 0 | 0.98 | 0.94 | 0.96 | 0.97 | 1.87USD |
| 1 | 0.99 | 0.97 | 0.98 | | |
| **Curie** | **Precision** | **Recall** | **F1** | **Accuracy** | **Cost** |
| 0 | 0.97 | 0.94 | 0.96 | 0.95 | 1.34USD |
| 1 | 0.95 | 0.96 | 0.95 | | |
| **Babbage** | **Precision** | **Recall** | **F1** | **Accuracy** | **Cost** |
| 0 | 0.93 | 0.91 | 0.92 | 0.95 | 0.74USD |
| 1 | 0.92 | 0.96 | 0.94 | | |
| **Ada** | **Precision** | **Recall** | **F1** | **Accuracy** | **Cost** |
| 0 | 0.89 | 0.90 | 0.89 | 0.90 | 0.13USD |
| 1 | 0.91 | 0.89 | 0.89 | | |
| **RL** | **Precision** | **Recall** | **F1** | **Accuracy** | |
| 0 | 0.91 | 0.93 | 0.92 | 0.94 | |
| 1 | 0.93 | 0.92 | 0.92 | | |

Considering that the performance of the GPT fine-tuned model improves with updates, it is declared that the GPT-3 used in this study was fine-tuned using the official model via API, with the testing date being Dec. 6, 2022.

Figure 8 showcases the text visualization capabilities of the RealEXP interpretability analysis tool applied to various user reviews. Each panel illustrates the decision-making process of the RL model in interpreting and classifying user feedback, with decision trees displaying key words and phrases that contributed to the model's predictions. The examples include negative reviews about hotel experiences, such as a superb breakfast, poor front desk service, and unsatisfactory accommodation. These visualizations highlight the tool's ability to identify critical factors influencing sentiment and provide interpretable insights into how the model arrives at its conclusions.

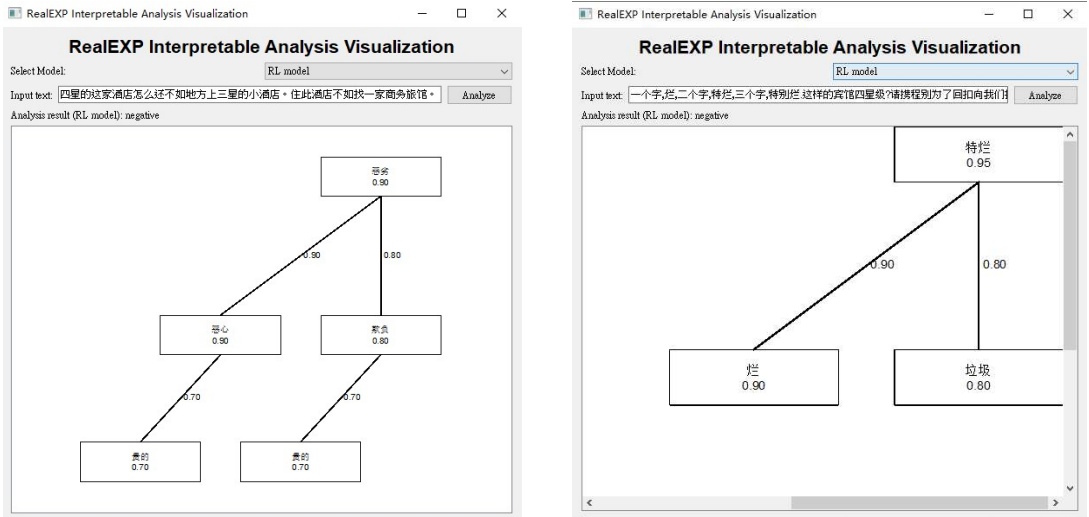

How can this four-star hotel be worse than a local three-star small hotel? Staying at this hotel is not as good as choosing a business inn.

One word: bad. Two words: very bad. Three words: extremely bad. A hotel like this, four stars? Please, Trip.com, stop recommending trash hotels just for kickbacks.

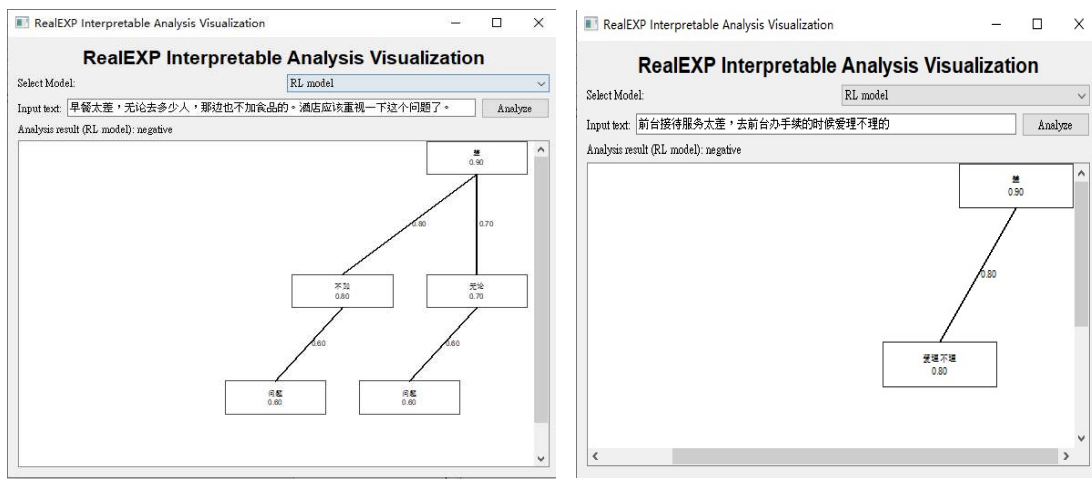

The breakfast is terrible. No matter how many people are there, they never replenish the food. The hotel really needs to address this issue.

The front desk service is terrible. When handling check-in procedures at the front desk, the staff seemed indifferent and unapproachable.

**Fig.8.** Text Visualization of RealExp

## Conclusion

This paper introduces a novel interpretability computation method, Real Explainer (RealExp), which decouples the Shapley Value into individual feature importance and feature correlation importance, enabling more precise interpretability analysis through feature similarity calculations. By considering both individual feature contributions and interactions among features, RealExp significantly enhances the accuracy and reliability of model explanations.

Furthermore, this paper proposes a new interpretability evaluation criterion that provides deeper insights into the decision paths of deep learning models, moving beyond traditional accuracy-based metrics. Case studies in image classification demonstrate how RealExp aids in selecting suitable pre-trained models, highlighting the practical limitations of existing architectures (e.g., Vision Transformer) and offering targeted improvement strategies. In text sentiment analysis, RealExp facilitated the design of a two-stage machine learning model that operates independently of pre-trained models yet surpasses GPT-ADA in performance.

Future work will focus on expanding the applications of interpretable machine learning in diverse domains of data processing and management, aiming to foster continued advancement and innovation in this field.